\PassOptionsToPackage{dvipsnames,table}{xcolor}
\documentclass{article} % For LaTeX2e
\usepackage{iclr2027_conference,times}
\usepackage[T1]{fontenc}

\usepackage{amsmath,amsfonts,bm}

\def\eqref#1{equation~\ref{#1}}
\def\1{\bm{1}}

\DeclareMathAlphabet{\mathsfit}{\encodingdefault}{\sfdefault}{m}{sl}
\SetMathAlphabet{\mathsfit}{bold}{\encodingdefault}{\sfdefault}{bx}{n}

\usepackage{xcolor}
\usepackage{hyperref}
\usepackage{url}
\usepackage{xspace}
\usepackage{amsmath, amsthm, amssymb}
\usepackage{enumitem}
\usepackage[capitalize,noabbrev,nameinlink]{cleveref}
\usepackage{graphicx}
\usepackage{booktabs} % for professional tables
\usepackage{multirow}
\usepackage{wrapfig}
\usepackage{placeins}
\usepackage{adjustbox}
\usepackage{rotating}
\usepackage{lscape}
\usepackage{tabularx, array, booktabs}
\usepackage{longtable}
\theoremstyle{plain}

\theoremstyle{definition}

\theoremstyle{remark}

\newcommand{\mypar}[1]{\textbf{#1}\hspace{4pt}}

\usepackage{acronym}
\acrodef{GBDT}[GBDT]{gradient boosting decision tree}
\acrodef{MLP}[MLP]{multi-layer perceptron}
\acrodef{TFM}[TFM]{tabular foundation model}
\acrodef{PFN}[PFN]{prior-data fitted networks}
\acrodef{ICL}[ICL]{in-context learning}
\acrodef{KD}[KD]{knowledge distillation}

\acrodef{NLP}[NLP]{natural language processing}
\acrodef{LLM}[LLM]{large language model}
\acrodef{CV}[CV]{computer vision}
\acrodefplural{KL}[KL]{Kullback-Leibler}
\acrodef{RNN}[RNN]{recurrent neural network}

\title{Distillation of Tabular Foundation Models\\into Efficient Predictors}

\author{
Minho Jeong$^{1}$\quad
Dooho Lee$^{1}$\quad
Jinmo Lee$^{1}$\quad
Jaemin Yoo$^{1}$
\\
$^1$Nums AI
}

\iclrfinalcopy % Show authors and omit review line numbers in the preprint.
\begin{document}
% Keep float spacing natural rather than stretching it to fill each page.
\raggedbottom

\maketitle
% Override the conference publication header set by \maketitle.
\lhead{Preprint}

\begin{abstract}
Tabular foundation models (TFMs) achieve strong predictive performance through in-context learning, yet repeatedly conditioning on labeled data makes inference expensive.
Knowledge distillation can reduce this cost by transferring their predictive ability to lightweight, dataset-specific students.
However, the dependence of TFM predictions on both a labeled context and a query introduces two design questions: how to construct teacher supervision and whether expanding query coverage improves distillation.
We examine these questions across two TFMs and both neural and tree-based students, and derive an effective distillation recipe.
The recipe uses the full labeled training set as teacher context and trains students solely on teacher predictions for observed and synthetic queries.
On TabArena, the resulting students outperform their supervised trained tuned-and-ensembled counterparts by 57--98 Elo points.
Applied unchanged to TALENT, the same recipe improves matched default students on 236--258 of 300 datasets and reduces median primary error by 4.0--6.4\%.
The distilled students also achieve median inference speedups of $3.0$--$21.6\times$ over their teachers, offering a practical trade-off between predictive performance and repeated inference cost.
Code is available at \href{https://github.com/nums-ai/TFM_Distillation}{link}.
\end{abstract}

\section{Introduction}
\label{sec:introduction}

Tabular machine learning has been dominated by supervised methods such as tree models~\citep{chen2016xgboost,ke2017lightgbm,prokhorenkova2018catboost}
and multi-layer perceptrons~\citep{holzmuller2024better,gorishniy2025tabm,gorishniy2026tabpack}.
These methods support fast inference once trained but require dataset-specific fitting and often extensive hyperparameter tuning.
Recently, \acp{TFM}~\citep{hollmann2025tabpfn,ma2025tabdpt,qu2025tabicl,qu2026tabiclv2} have changed this paradigm with strong out-of-the-box performance without per-dataset updates.
The central idea of \acp{TFM} is \emph{in-context learning,} which enables zero-shot inference by conditioning predictions on labeled samples in the new dataset, not on the training samples observed earlier.

However, the in-context learning mechanism of \acp{TFM} does not eliminate the computational costs but rather shifts them from training to inference because the context must be processed every time a prediction is made.
Repeated TFM inference is therefore much more costly than using a trained supervised method.
Several recent works reduce this burden through context
selection~\citep{thomas2024retrieval,liu2025tabpfn},
routing queries to specialized local contexts~\citep{xu2025mixture},
compressing contexts into latent
representations~\citep{zabergja2026taco},
or caching reusable context
representations~\citep{grinsztajn2025tabpfnv2.5,qu2026tabiclv2}.
These approaches improve the overall efficiency of inference but retain context-dependent computation at serving time.

\Ac{KD}~\citep{bucilua2006model,hinton2015distilling} gives a complementary
approach that transfers the predictive ability of a \ac{TFM} to a lightweight, dataset-specific student.
Once created, a distilled student supports instant inference as many times as we want, saving huge computational costs.
However, because of the unique properties of \acp{TFM} as in-context learners, distilling a \ac{TFM} introduces two key essential design choices: \emph{(i) supervision construction} asks how teacher supervision should be formed from training data, while \emph{(ii) query coverage} asks whether and how expanding teacher queries improves distillation.

\emph{Supervision construction.}
What we distill from a \ac{TFM} to a student is not its full capacity; we distill its predictive power conditioned on a specific labeled context.
For every labeled observation in the training data, we can either (i) include it in the context to make the TFM's predictions more focused around the data manifold, or (ii) exclude it from the context but include it in the query data, possibly providing its true label as an additional supervised signal.
We find that students perform best when trained solely on teacher predictions generated using the full labeled training set as context.
Under this design, TFM supervision can fully replace direct label supervision.

\emph{Query coverage.}
The second choice concerns which inputs are used to obtain teacher supervision.
One way to make full use of the teacher is to expand query coverage beyond the observed training rows.
By querying synthetic inputs, we can obtain additional supervision for the student even without collecting new labels.
We test whether this expansion improves distillation using four simple, task-agnostic query generators that perturb or recombine observed features.
All four generators improve aggregate performance for every teacher--student pair.
These results show that useful teacher supervision can extend beyond observed rows even with simple query generation.

\begin{figure}[!t]
    \centering
    \includegraphics[width=\linewidth]{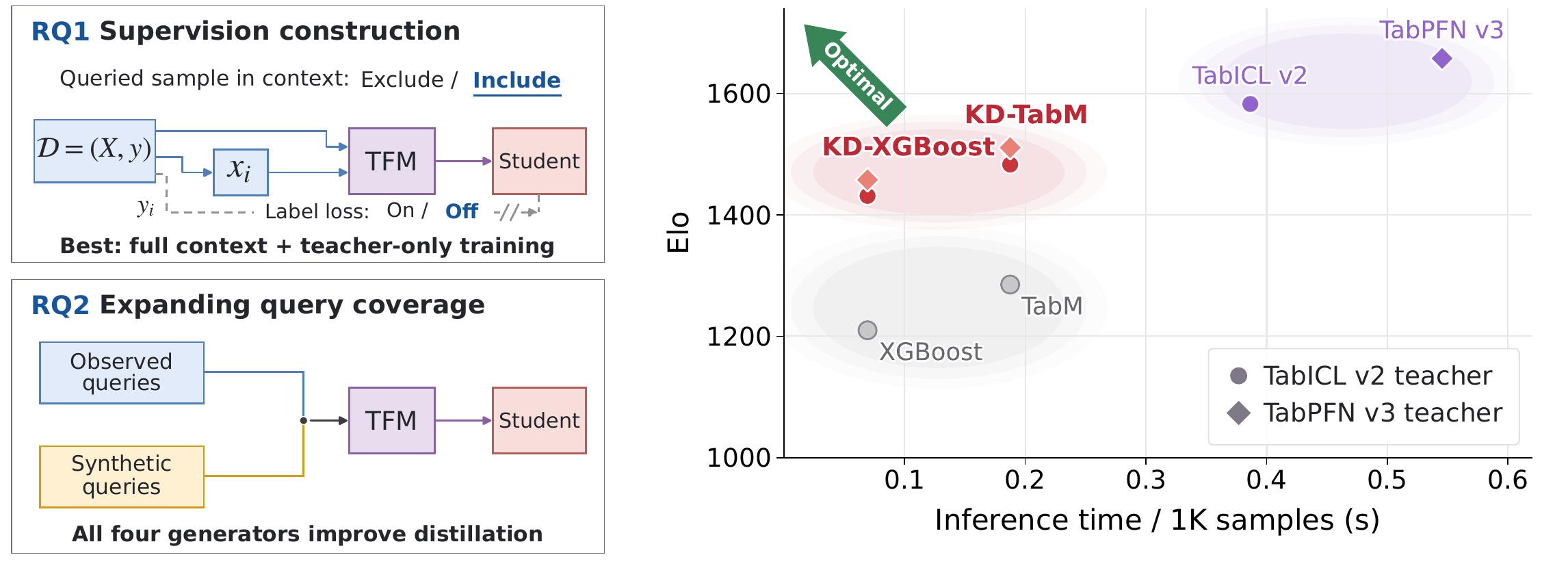}
\caption{
    \textbf{Constructing supervision for TFM distillation.}
    (Left) We study how to construct teacher supervision for ditillation (RQ1) and whether and how expanding teacher queries improves distillation (RQ2).
    We find that training solely on teacher predictions generated with the full labeled training set as context achieves the best performance.
    % , and all four tested query generators improve both student families.
    (Right) Elo versus median inference time per 1K samples on TabArena for default lightweight students, their distilled counterparts, and the TFM teachers, with Elo calibrated jointly over the same 76 configurations as in Section~\ref{sec:evaluation}.
    }
    \label{fig:intro}
\end{figure}

Figure~\ref{fig:intro} summarizes these design choices
and illustrates the resulting trade-off between predictive
performance and inference cost on TabArena.

Together, these findings yield a simple but effective distillation recipe of \acp{TFM}: use the full labeled training set as teacher context and train the student solely on teacher predictions for observed and synthetic queries.
Our distilled students with single default student configurations outperform tuned-and-ensembled students on TabArena~\citep{erickson2025tabarena} by 57--98 Elo points.
Moreover, the same recipe improves students across binary classification, multiclass classification, and regression on 300 TALENT datasets~\citep{liu2025talent}.
The distilled students trade a one-time distillation cost and some accuracy relative to their teachers for faster repeated inference, achieving median speedups of $3.0$--$21.6\times$, with the inference advantage growing as the teacher context size increases.

\section{Related Work}
\label{sec:related-work}

\paragraph{Tabular foundation models.}

\Acp{PFN} learn to approximate posterior predictions by pretraining over
synthetic tasks and conditioning on labeled samples at inference
time~\citep{mullertransformers}.
TabPFN introduced this paradigm to tabular prediction~\citep{
hollmann2023tabpfn,hollmann2025tabpfn}, and recent \acp{TFM} have
substantially improved predictive performance and scalability
~\citep{qu2025tabicl,qu2026tabiclv2,grinsztajn2025tabpfnv2.5, grinsztajn2026tabpfnv3, cho2026causilotechnicalreport}.
Unlike supervised models, which fit a separate predictor to each dataset, they adapt through in-context learning: predictions are created by conditioning on labeled examples from the target dataset without dataset-specific parameter updates.
This enables strong out-of-the-box prediction, but it also means that inference continues to depend on the labeled context even after the model is fully deployed.

\paragraph{Improving context efficiency.}

The context dependence of \acp{TFM} has motivated several approaches to improve efficiency.
One direction improves the efficiency of the context itself.
LoCalPFN constructs local contexts through nearest-neighbor
retrieval~\citep{thomas2024retrieval}, while MixturePFN routes queries
to specialized experts with cluster-specific contexts~\citep{xu2025mixture}.
BETA combines lightweight encoder adaptation with bootstrapped
sampling to scale contextual prediction~\citep{liu2025tabpfn}.
TACO jointly trains a context compressor and a predictor, enabling
predictions to condition on a compact latent representation of the
training data~\citep{zabergja2026taco}.
TabPFN-2.5 and TabICLv2 accelerate repeated inference by caching
reusable context representations~\citep{grinsztajn2025tabpfnv2.5,qu2026tabiclv2}.
These approaches reduce inference cost, but predictions still depend
on context examples or their cached representations at serving time.

\paragraph{Knowledge distillation of \acp{TFM}.}

A complementary direction transfers the information of a \ac{TFM} into a
dataset-specific predictor that can be served independently.
MotherNet generates the weights of a compact neural network from a
dataset~\citep{muller2025mothernet}, while knowledge distillation
trains a lightweight student from teacher predictions.
Classical distillation may combine teacher predictions with ground-truth
labels~\citep{bucilua2006model,hinton2015distilling}.
For tabular models, FAST-DAD distills ensemble teachers into individual
predictors using observed labels together with teacher predictions on
synthetic inputs generated by Gibbs sampling.~\citep{fakoor2020fast}.
Recent TFM distillation methods likewise instantiate particular choices of
teacher supervision: Pocket Foundation Models combines out-of-fold teacher
predictions with direct label supervision~\citep{tanna2026pocket}.
In concurrent work, GEAR combines distillation on synthetic samples with
sophiscated generators and
that combines direct label supervision and out-of-fold teacher
supervision~\citep{qin2026gear}.

\section{Designing TFM Distillation}
\label{sec:design}

We study two design choices for distilling a pretrained \ac{TFM} into a lightweight, dataset-specific student.
The first, \emph{supervision construction}, concerns how teacher predictions are generated for training datas and whether they are combined with direct label supervision.
The second, \emph{query coverage}, concerns whether and how querying additional inputs improves distillation.

\mypar{Experimental setup.}
Our design studies use 51 TabArena datasets with three official outer splits.
We consider all four teacher--student pairs formed by two strong TFM teachers, TabICLv2~\citep{qu2026tabiclv2} and TabPFN v3~\citep{grinsztajn2026tabpfnv3}, and two structurally distinct students, the neural model TabM~\citep{gorishniy2025tabm} and the tree-based model XGBoost~\citep{chen2016xgboost}.
Our experiments build on the official TabArena training and evaluation pipeline, retaining its default student configurations, preprocessing, and protocol.
Configurations are compared using aggregate TabArena Elo, calibrated jointly within each comparison.
Further details appear in Appendix~\ref{app:sec3-experimental-details}.

\subsection{Supervision Construction}
\label{sec:target_routing}

Let $\mathcal{D}=\{(x_i,y_i)\}_{i=1}^{n}$ denote the labeled training dataset.
A TFM prediction $q_T(\cdot \mid x,C)$ depends on a labeled context $C$ and a query input $x$.
The context provides evidence for prediction, while the output for the query serves as supervision for the student.
When querying a training data, we can either exclude it from the context or include it together with its label.

We compare two protocols that generate teacher targets for the same samples.
Out-of-fold (OOF) prediction queries each held-out fold using the remaining folds as context, excluding the queried sample and its label.
Full-context (FC) prediction queries each training sample using the full labeled training set $\mathcal{D}$ as context, including the queried sample and its label.
Let $\{\mathcal{D}_k\}_{k=1}^{K}$ partition $\mathcal{D}$, and let $f(i)$ denote the fold containing $(x_i, y_i)$:
\begin{equation}
q_i^{\mathrm{OOF}}
=
q_T\!\left(
\cdot \mid x_i,
\mathcal{D}\setminus\mathcal{D}_{f(i)}
\right),
\qquad
q_i^{\mathrm{FC}}
=
q_T(\cdot \mid x_i,\mathcal{D}).
\label{eq:teacher_targets}
\end{equation}
Either set of teacher targets can be further combined with direct label supervision through the objective
\begin{equation}
\mathcal{L}(\theta)
=
\frac{1}{n}\sum_{i=1}^{n}
\left[
\alpha\,
\ell_{\mathrm{KD}}
\!\left(
q_{S_\theta}(\cdot \mid x_i), q_i
\right)
+
(1-\alpha)\,
\ell_{\mathrm{sup}}
\!\left(
q_{S_\theta}(\cdot \mid x_i), y_i
\right)
\right],
\label{eq:routing_loss}
\end{equation}
where $q_{S_\theta}$ denotes the student prediction and $q_i$ is either $q_i^{\mathrm{OOF}}$ or $q_i^{\mathrm{FC}}$.
The losses $\ell_{\mathrm{KD}}$ and $\ell_{\mathrm{sup}}$ measure agreement with the teacher target and the ground-truth label, respectively.
Thus, $\alpha=1$ uses only teacher predictions, $\alpha=0$ uses only ground-truth labels, and intermediate values combine the two sources of supervision.

\begin{figure}[!htbp]
    \centering
    \includegraphics[width=\linewidth]{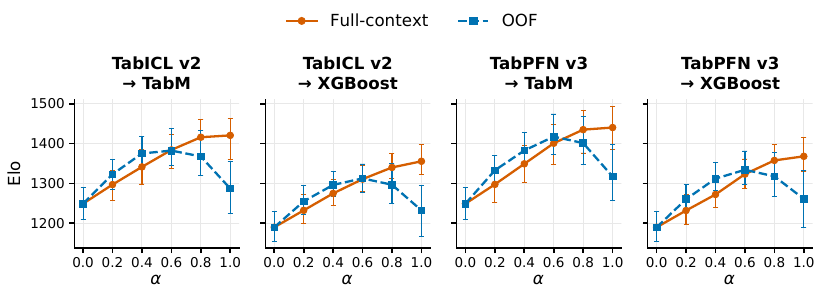}
    \caption{\textbf{Teacher-only supervision with the full labeled context achieves the best aggregate performance.}
    Full-context targets reach the highest Elo at \(\alpha=1\) in every teacher--student pair, exceeding the best OOF--label mixture at \(\alpha=0.6\).}
    \label{fig:rq1}
\end{figure}

\mypar{Full-context teacher predictions provide the best supervision.}
Across all four teacher--student pairs, the best tested configuration trains the student solely on teacher predictions generated with the full labeled training set as context.
We sweep $\alpha \in \{0, 0.2, 0.4, 0.6, 0.8, 1\}$ for both protocols.
Figure~\ref{fig:rq1} shows that full-context targets achieve their highest aggregate Elo at $\alpha=1$, with no direct label loss.
In contrast, OOF targets benefit from adding direct label supervision, with Elo peaking at $\alpha=0.6$ in every pair.
Full-context teacher-only supervision exceeds the best tested OOF--label mixture by 22.8--43.0 Elo points.

\mypar{Finding 1.}
\textbf{Teacher predictions generated with the full labeled training set as context can replace direct label supervision.}
For both teachers and student families studied, learning solely from these predictions yields the best aggregate performance among the tested supervision configurations.

We therefore fix the full labeled training set as teacher context and use teacher-only supervision ($\alpha=1$) for the remaining experiments.

\subsection{Expanding Query Coverage}

One simple way to make fuller use of the teacher is to expand query coverage beyond the real training rows.
By querying synthetic inputs, we can obtain additional supervision without collecting new labels.
We ask whether this expansion improves distillation when the queries are generated by simple perturbations or recombinations of real sample's features.
For each synthetic input $\tilde{x}$, we obtain a teacher target using
\begin{equation}
q^{\mathrm{syn}}
=
q_T(\cdot \mid \tilde{x}, \mathcal{D}),
\end{equation}
where the labeled context $\mathcal{D}$ remains the same as for real queries.
The student then learns from teacher predictions on both real and synthetic queries.

\mypar{Four simple query generators.}
We evaluate four task-agnostic, non-parametric generators at a common budget of one synthetic query per real row.
\textsc{CutMix} independently selects each feature from one of two sampled rows with equal probability.
\textsc{Mixup} interpolates numerical features between two rows using feature-wise mixing coefficients, while categorical features select between the corresponding values.
\textsc{FullMix} independently draws each feature from a randomly selected real row.
\textsc{Perturb} adds standard-deviation-scaled Gaussian noise to numerical features and resamples categorical values from their empirical marginals.

\begin{figure}[!htbp]
    \centering
    \includegraphics[width=\linewidth]{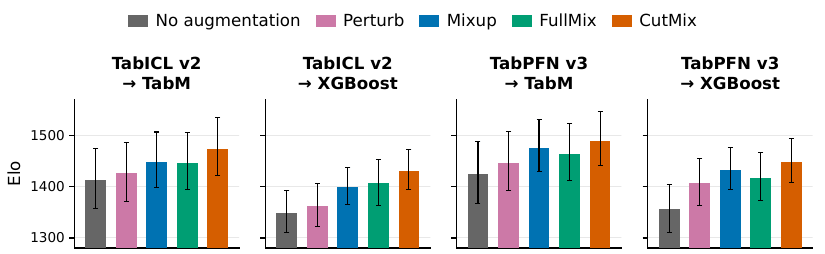}
    \vspace{-2mm}
    \caption{\textbf{Simple synthetic queries improve distillation across all four teacher--student pairs.}
    Each of the four query generators yields higher Elo than distillation using only observed queries, with \textsc{CutMix} achieving the highest Elo in every pair.}
    \label{fig:rq2}
    \vspace{-2mm}
\end{figure}

\mypar{Expanding query coverage improves distillation.}
Figure~\ref{fig:rq2} compares distillation using only real queries with adding synthetic queries from each generator.
All four yield higher Elo for every teacher--student pair, with gains ranging from 14 to 91 Elo points.
\textsc{CutMix} achieves the highest Elo in all four pairs, and each of the other generators also improves over real queries alone.

\mypar{Finding 2.}
\textbf{Simple synthetic queries extend useful teacher supervision beyond real training rows.}
The gains hold across all four tested generators, both teachers, and both students.

\begin{figure}[!htbp]
    \centering
    \includegraphics[width=\linewidth]{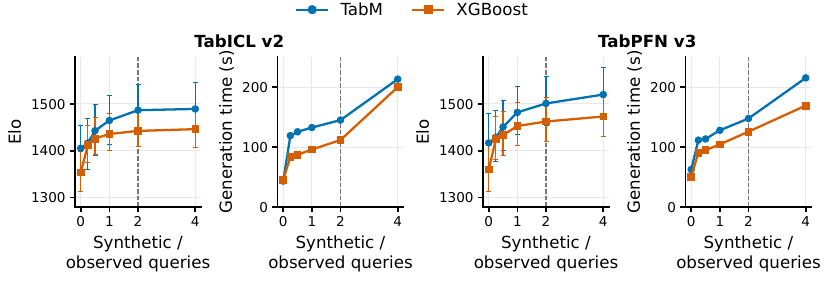}
    \vspace{-2mm}
    \caption{ \textbf{Effect of the synthetic-query budget for \textsc{CutMix}.}
    Each teacher group pairs mean Elo with teacher prediction generation time.
    Increasing the synthetic-to-real query ratio improves mean Elo across both teachers and student families while increasing generation time.
    The dashed line marks the $2{:}1$ ratio used in our fixed recipe, chosen as a common operating point on this performance--generation-cost trade-off rather than tuned per dataset.
    }
    \label{fig:rq2_ratio}
    \vspace{-2mm}
\end{figure}

\mypar{More synthetic queries provide further gains at additional teacher cost.}
Using \textsc{CutMix}, Figure~\ref{fig:rq2_ratio} jointly tracks Elo and teacher prediction generation time as the synthetic-to-real query ratio increases.
Increasing the ratio from $1{:}1$ to $2{:}1$ adds $6.5$--$21.9$ Elo across the four pairs, while mean generation time increases by $9$--$20\%$.
The benefit of a larger budget varies across pairs: moving from $2{:}1$ to $4{:}1$ adds only $2.7$ and $3.8$ Elo for TabICLv2--TabM and TabICLv2--XGBoost, respectively, at $47\%$ and $79\%$ higher mean generation cost.
TabPFN v3--TabM continues to gain $19.0$ Elo over the same increase, with $46\%$ more generation time.

Although $4{:}1$ achieves the highest Elo point estimate in every pair, $2{:}1$ already improves substantially over real queries alone while requiring less target-generation time.
We therefore use $2{:}1$ as a common performance--cost compromise for the remaining experiments.

\mypar{Final recipe.}
\textbf{Together, these studies define a single distillation recipe that we fix for all subsequent experiments.}
We use the full labeled training set as teacher context and train the student solely on teacher predictions and \textsc{CutMix} synthetic queries at a $2{:}1$ synthetic-to-real ratio.
\section{Evaluation}
\label{sec:evaluation}

Section~\ref{sec:design} fixes a single recipe for \ac{TFM} distillation.
We now evaluate this recipe on TabArena~\citep{erickson2025tabarena} and TALENT~\citep{liu2025talent}.

\mypar{Experimental setup.}
As in Section~\ref{sec:design}, we use
TabICLv2~\citep{qu2026tabiclv2} and TabPFN v3~\citep{grinsztajn2026tabpfnv3}
as teachers, and TabM~\citep{gorishniy2025tabm} and
XGBoost~\citep{chen2016xgboost} as students.
For KD, we'll use the selected recipe in
Section~\ref{sec:design} without further reselection and use the default student
configurations throughout.
On TabArena~\citep{erickson2025tabarena}, we evaluate on all 51 datasets under
the official benchmark protocol.
We compute aggregate TabArena scores jointly over the 72 official baseline configurations and our four KD configurations, using default student configurations and a $2{:}1$ synthetic-to-observed query ratio ($r=2$).
On TALENT~\citep{liu2025talent}, we evaluate on all 300 datasets using the
provided train/validation/test splits and five training seeds.
We measure error by $1-\mathrm{ROC\mbox{-}AUC}$ for binary classification,
log loss for multiclass classification, and RMSE for regression.
For each dataset, we first average errors over splits or seeds and then compute
matched relative-error reductions.
Aggregate results weight datasets equally, with uncertainty estimated by paired
dataset-bootstrap 95\% confidence intervals.
Appendix~\ref{app:sec4-experimental-details} provides full protocols, statistical details,
and additional analyses.

\mypar{Baselines.}
On both benchmarks, we compare KD against architecture-matched default students
trained directly without distillation.
On TabArena, we additionally compare against tuned and tuned-and-ensembled
versions of the same students, all 15 conventional supervised model families in
the benchmark pool, including RealMLP~\citep{holzmuller2024better},
LightGBM~\citep{ke2017lightgbm}, and CatBoost~\citep{prokhorenkova2018catboost},
as well as the teacher models.

\subsection{Distillation Improves the Same Student Beyond Direct Optimization}
\label{sec:eval_tabarena}

% Elo source: artifacts/tabarena_76/paired_elo_contrasts.csv.
% Full split scope; error-based KD gains and W/T/L are independent of the comparison pool.
\begin{table}[t]
\centering
\small
\caption{\textbf{KD improves each student beyond supervised tuning and ensembling.}
Each KD model is compared with the tuned-and-ensembled version of the same architecture on 51 TabArena datasets.
KD gain is the median relative primary-error reduction, W/T/L counts dataset-level wins, ties, and losses.
Elo is calibrated jointly over all 76 configurations; differences are computed before rounding.
Brackets show paired dataset-bootstrap 95\% intervals.}
\label{tab:section4-tabarena-paired-main}
\setlength{\tabcolsep}{5pt}
\renewcommand{\arraystretch}{1.08}
\begin{tabularx}{\linewidth}{@{}Xrrr@{}}
\toprule
Model & $\Delta$Elo $\uparrow$ & KD gain (\%) $\uparrow$ & W/T/L \\
\midrule
TabICL v2 $\rightarrow$ TabM & +56.7 [3.9, 107.4] & 1.30 [0.61, 1.93] & 33/0/18 \\
TabICL v2 $\rightarrow$ XGBoost & +71.4 [14.3, 128.7] & 1.69 [0.38, 2.21] & 35/0/16 \\
\addlinespace[2pt]
TabPFN v3 $\rightarrow$ TabM & +84.5 [34.1, 135.1] & 1.38 [0.71, 2.36] & 36/0/15 \\
TabPFN v3 $\rightarrow$ XGBoost & +97.9 [41.9, 156.9] & 1.91 [0.94, 3.15] & 37/0/14 \\
\bottomrule
\end{tabularx}
\vspace{3pt}
\end{table}

KD outperforms tuned-and-ensembled direct training for both student
architectures.
Across the four teacher--student pairs, distilled students gain
$56.7$--$97.9$ Elo over the matched tuned ensemble, with all paired 95\%
confidence intervals above zero.
They also win on $33$--$37$ of 51 datasets, with median relative-error
reductions of $1.30$--$1.91\%$ and positive confidence intervals for every pair.

% Source: artifacts/tabarena_76/leaderboard.csv; generated by scripts/update_tabarena_76_manuscript.py.
\begin{table}[t]
\centering
\small
\caption{\textbf{Distilled students and leading supervised models on TabArena.}
Results cover 51 datasets and 816 dataset--split pairs using a joint pool of 72 official baseline configurations and four KD configurations.
Win rate averages pairwise outcomes against the other 75 methods in this pool.
Distilled students use default student configurations and $r=2$, with no student hyperparameter search.
Light-blue row shading identifies distilled students.
Among the displayed non-teacher models, \textcolor{Red}{\textbf{1st}}, \textcolor{Orange}{\textbf{2nd}}, and \textcolor{Emerald}{\textbf{3rd}} best results in each column are highlighted.
Teachers are reported separately as references and excluded from these markings.}
\label{tab:section4-tabarena-supervised}
\label{tab:final-recipe}
\label{tab:tabarena_main}
\setlength{\tabcolsep}{5pt}
\renewcommand{\arraystretch}{1.08}
\begin{tabularx}{\linewidth}{@{}Xrrr@{}}
\toprule
Model & Elo $\uparrow$ & Improvability (\%) $\downarrow$ & Win rate (\%) $\uparrow$ \\
\midrule
\rowcolor{blue!6}
TabPFN v3 $\rightarrow$ TabM & \textcolor{Red}{\textbf{1510.7}} & \textcolor{Orange}{\textbf{11.21}} & \textcolor{Red}{\textbf{76.4}} \\
RealMLP (Tuned+Ensemble) & \textcolor{Orange}{\textbf{1489.9}} & \textcolor{Red}{\textbf{11.09}} & \textcolor{Orange}{\textbf{74.5}} \\
\rowcolor{blue!6}
TabICL v2 $\rightarrow$ TabM & \textcolor{Emerald}{\textbf{1483.0}} & \textcolor{Emerald}{\textbf{11.44}} & \textcolor{Emerald}{\textbf{73.9}} \\
\rowcolor{blue!6}
TabPFN v3 $\rightarrow$ XGBoost & 1457.8 & 12.41 & 71.4 \\
\rowcolor{blue!6}
TabICL v2 $\rightarrow$ XGBoost & 1431.3 & 12.62 & 68.8 \\
TabM (Tuned+Ensemble) & 1426.2 & 12.50 & 68.3 \\
LightGBM (Tuned+Ensemble) & 1412.5 & 13.49 & 66.8 \\
CatBoost (Tuned+Ensemble) & 1398.1 & 13.10 & 65.3 \\
TabM (Tuned) & 1373.7 & 13.36 & 62.7 \\
XGBoost (Tuned+Ensemble) & 1359.9 & 14.26 & 61.2 \\
XGBoost (Tuned) & 1338.1 & 14.57 & 58.7 \\
TabM (Default) & 1285.2 & 15.57 & 52.7 \\
XGBoost (Default) & 1209.8 & 17.35 & 44.2 \\
\midrule
\multicolumn{4}{@{}l}{\textit{Teacher models}} \\
\addlinespace[2pt]
TabPFN v3 & 1657.8 & 6.83 & 87.2 \\
TabICL v2 & 1582.7 & 7.60 & 82.2 \\
\bottomrule
\end{tabularx}
\vspace{3pt}
\end{table}

The distilled students also rank strongly against the broader TabArena pool.
All four outperform the strongest configuration of 14 of the 15 conventional
supervised model families.
RealMLP is the only exception, although TabPFN v3--TabM also exceeds its tuned and
ensembled variant.
Table~\ref{tab:section4-tabarena-supervised} reports selected baselines; the full
comparison is in Appendix~\ref{app:full_results}.

\subsection{Teacher Advantage Transfers through KD}
\label{sec:eval_regimes}

To assess whether KD actually transfers the teacher's predictive advantage,
we measure how much of the teacher--student error gap is recovered by the
distilled student and whether larger teacher advantages translate into larger KD gains.
Across datasets where the teacher reduces raw-student error by more than $1\%$,
KD recovers a median $67.9$--$70.9\%$ of the teacher--student error gap
(Table~\ref{tab:section4-transfer-summary}).
Thus, when the TFM is meaningfully stronger than the student, roughly two-thirds
of its predictive advantage is retained in the lightweight student.

\begin{figure}[t]
\centering
\includegraphics[width=\linewidth]{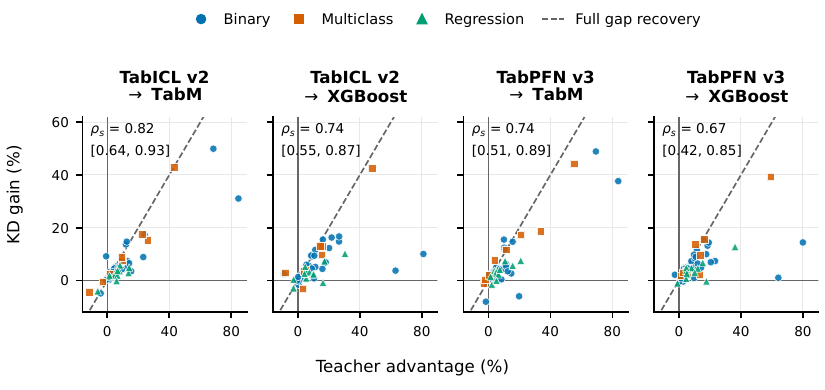}
\caption{\textbf{Transfer of teacher gains to distilled students across 51 TabArena datasets.}
Teacher and KD gains are measured as relative error reduction over the raw student,
$100(1-E/E_{\rm raw})$; the dashed line denotes full recovery of the teacher gain.
Annotations report Spearman correlations with 95\% bootstrap CIs.
Because both axes share the raw baseline, these correlations are descriptive.}
\label{fig:section4-knowledge-transfer}
\end{figure}

\begin{table}[t]
\centering
\small
\caption{\textbf{Recovery of teacher gains on TabArena.}
We restrict to datasets where the teacher reduces default-student test error by more than $1\%$.
Gap recovery is $100(E_{\rm raw}-E_{\rm KD})/(E_{\rm raw}-E_T)$, using errors averaged over official splits; entries report the dataset median with bootstrap 95\% CIs, and W/T/L against the raw student.}
\label{tab:section4-transfer-summary}
\setlength{\tabcolsep}{5pt}
\renewcommand{\arraystretch}{1.08}
\begin{tabularx}{\linewidth}{@{}Xrrr@{}}
\toprule
Model & $n$ & Gap recovery (\%) & KD W/T/L \\
\midrule
TabICL v2 $\rightarrow$ TabM & 43 & 70.5 [57.4, 90.0] & 42/0/1 \\
TabICL v2 $\rightarrow$ XGBoost & 43 & 69.4 [54.0, 80.6] & 41/0/2 \\
\addlinespace[2pt]
TabPFN v3 $\rightarrow$ TabM & 46 & 70.9 [55.3, 91.6] & 43/0/3 \\
TabPFN v3 $\rightarrow$ XGBoost & 48 & 67.9 [52.7, 78.1] & 46/0/2 \\
\bottomrule
\end{tabularx}
\vspace{3pt}
\end{table}

This transfer is also reflected at the dataset level.
Larger teacher advantages are associated with larger KD gains
(Figure~\ref{fig:section4-knowledge-transfer}), with Spearman correlations of
$0.67$--$0.82$ across the four teacher--student pairs and all bootstrap 95\%
confidence intervals above zero.
This relationship indicates that the benefit of KD closely tracks the
dataset-specific advantage provided by the teacher.
Consistent with this pattern, when the teacher outperforms the raw student,
KD also improves the student on nearly all such datasets, with only one to three
exceptions per pair.

\subsection{One-Time Distillation Cost vs.\ Inference Efficiency}
\label{sec:eval_efficiency}

We finally examine whether the one-time cost of distillation can be amortized
through faster repeated inference, and in which serving regimes this trade-off is
most favorable.
Table~\ref{tab:section4-runtime-tradeoff} quantifies the deployment trade-off
across all 51 TabArena datasets using 32-row requests on the same H200 host with
32 pinned CPU cores.
Serving measurements start from resident models and exclude initial loading,
whereas offline cost includes teacher-target generation, student fitting, and
associated overhead (Appendix~\ref{app:section4-timing}).
Across the four teacher--student pairs, dataset-paired median inference speedups
range from $3.0$ to $21.6\times$, while the distilled students incur median
relative primary-error increases of $1.72$--$2.76\%$ and median one-time costs of
$48.0$--$192.4$ seconds.

\begin{table}[t]
\centering
\small
\caption{\textbf{Inference savings, prediction error, and distillation cost.}
Results on 51 TabArena datasets at outer split 0.
Latencies are per 32-row request; speedup uses dataset-paired teacher/KD ratios, and $\Delta$Error reports the relative error increase with paired-bootstrap 95\% CIs.
Offline time indicates end-to-end distillation tuime.
Point estimates are dataset medians and brackets denote IQRs.}
\label{tab:section4-runtime-tradeoff}
\setlength{\tabcolsep}{5pt}
\renewcommand{\arraystretch}{1.08}
\begin{tabularx}{\linewidth}{@{}Xrrrrr@{}}
\toprule
Model & Teacher (ms) & KD (ms) & Speedup & $\Delta$Error (\%) & Offline (s) \\
\midrule
TabICL v2 $\rightarrow$ TabM & 232.2 & 47.8 & $3.0\times$ & \shortstack[r]{1.72\\{\scriptsize [0.53, 2.94]}} & \shortstack[r]{186.3\\{\scriptsize [93.1, 450.9]}} \\
TabICL v2 $\rightarrow$ XGBoost & 232.2 & 29.1 & $8.7\times$ & \shortstack[r]{2.33\\{\scriptsize [1.07, 5.54]}} & \shortstack[r]{48.0\\{\scriptsize [20.0, 146.5]}} \\
\addlinespace[2pt]
TabPFN v3 $\rightarrow$ TabM & 566.9 & 46.7 & $11.1\times$ & \shortstack[r]{2.76\\{\scriptsize [0.70, 4.12]}} & \shortstack[r]{192.4\\{\scriptsize [110.4, 563.4]}} \\
TabPFN v3 $\rightarrow$ XGBoost & 566.9 & 29.7 & $21.6\times$ & \shortstack[r]{2.53\\{\scriptsize [1.33, 4.17]}} & \shortstack[r]{59.6\\{\scriptsize [30.6, 173.9]}} \\
\bottomrule
\end{tabularx}
\vspace{3pt}
\end{table}

To relate this upfront cost to repeated serving, let $A$ denote the one-time
distillation cost and $t_T$ and $t_S$ the teacher and student latencies per
32-row request, measured in the same time units.
We estimate the number of requests required to amortize distillation as
\begin{equation}
N_{\text{break-even}}=\frac{A}{t_T-t_S}.
\label{eq:section4-break-even-requests}
\end{equation}
If $t_T\leq t_S$, no finite break-even point exists and we set
$N_{\text{break-even}}=\infty$.
Table~\ref{tab:section4-break-even-requests} groups datasets by teacher context
size and reports medians of the dataset-level speedups and break-even request
counts within each group.

% Generated by analysis/section4/break_even_requests.py from saved dataset-level timings.
\begin{table}[t]
\centering
\small
\caption{\textbf{Inference savings and break-even requests by teacher context size.}
Entries report dataset medians of inference speedup $t_T/t_S$ and break-even 32-row requests.
All 51 datasets are included; students are faster on all 15 datasets with at least 10K teacher context rows.}
\label{tab:section4-break-even-requests}
\setlength{\tabcolsep}{4pt}
\renewcommand{\arraystretch}{1.08}
\begin{tabularx}{\linewidth}{@{}Xrrrrrr@{}}
\toprule
Context rows & \multicolumn{2}{c}{$<1$K ($n=10$)} & \multicolumn{2}{c}{$1$K--$<10$K ($n=26$)} & \multicolumn{2}{c}{$\geq10$K ($n=15$)} \\
\cmidrule(lr){2-3}\cmidrule(lr){4-5}\cmidrule(l){6-7}
Model & Speedup & $N_{\text{break-even}}$ & Speedup & $N_{\text{break-even}}$ & Speedup & $N_{\text{break-even}}$ \\
\midrule
TabICL v2 $\rightarrow$ TabM & 1.9$\times$ & 3,260.5 & 2.1$\times$ & 2,073.6 & 31.4$\times$ & 468.3 \\
TabICL v2 $\rightarrow$ XGBoost & 3.0$\times$ & 393.0 & 6.8$\times$ & 211.2 & 23.9$\times$ & 160.1 \\
\addlinespace[2pt]
TabPFN v3 $\rightarrow$ TabM & 10.2$\times$ & 503.2 & 6.8$\times$ & 388.9 & 53.7$\times$ & 266.5 \\
TabPFN v3 $\rightarrow$ XGBoost & 13.1$\times$ & 97.5 & 20.3$\times$ & 97.6 & 32.3$\times$ & 100.8 \\
\bottomrule
\end{tabularx}
\vspace{3pt}
\end{table}

The deployment benefit grows markedly with teacher context size.
With at least 10K context rows, the distilled student is faster on all 15
datasets for every teacher--student pair, with median speedups of
$23.9$--$53.7\times$, compared with $1.9$--$13.1\times$ below 1K rows.
For TabICLv2--TabM, the larger serving advantage also reduces the median
break-even volume from about $3{,}261$ to $468$ requests.
However, speedup and amortization are not equivalent: the break-even point also
depends on the upfront distillation cost and the absolute latency saved per request.
For example, for TabPFN v3--XGBoost, the median speedup rises from
$13.1\times$ to $32.3\times$, while the median break-even volume remains near
100 requests.
Overall, distillation is most compelling when the teacher incurs substantial
context-dependent inference cost and the resulting student is reused for repeated
prediction.

\subsection{Evaluation on TALENT}
\label{sec:eval_talent}

\begin{table}[t]
\centering
\small
\caption{\textbf{The fixed recipe improves both students across TALENT task types.}
KD gains are median relative primary-error reductions against the same default supervised student, after averaging five matched seeds.
Brackets show paired dataset-bootstrap 95\% intervals.
W/T/L counts wins, ties, and losses on all 300 datasets; gain summaries exclude three zero-error baselines.}
\label{tab:section4-talent_summary}
\setlength{\tabcolsep}{5pt}
\renewcommand{\arraystretch}{1.08}
\begin{tabularx}{\linewidth}{@{}Xrrrrr@{}}
\toprule
 & \multicolumn{4}{c}{Median KD gain (\%) $\uparrow$} & \\
\cmidrule(lr){2-5}
Model & \textbf{All} & Binary & Multiclass & Regression & W/T/L \\
 & {\scriptsize $n=297$} & {\scriptsize $n=117$} & {\scriptsize $n=80$} & {\scriptsize $n=100$} & {\scriptsize $n=300$} \\
\midrule
\multirow{2}{*}{TabICL v2 $\rightarrow$ TabM} & \textbf{4.20} & 8.46 & 5.80 & 1.44 & \multirow{2}{*}{238/3/59} \\
 & {\scriptsize [3.14, 5.33]} & {\scriptsize [5.55, 10.41]} & {\scriptsize [2.81, 10.65]} & {\scriptsize [0.47, 2.99]} & \\
\addlinespace[2pt]
\multirow{2}{*}{TabICL v2 $\rightarrow$ XGBoost} & \textbf{5.69} & 9.53 & 9.08 & 2.71 & \multirow{2}{*}{253/2/45} \\
 & {\scriptsize [4.60, 7.50]} & {\scriptsize [7.32, 12.23]} & {\scriptsize [7.21, 13.46]} & {\scriptsize [1.74, 3.49]} & \\
\addlinespace[4pt]
\multirow{2}{*}{TabPFN v3 $\rightarrow$ TabM} & \textbf{4.02} & 7.90 & 7.63 & 1.66 & \multirow{2}{*}{236/3/61} \\
 & {\scriptsize [3.13, 6.15]} & {\scriptsize [5.00, 9.98]} & {\scriptsize [3.36, 10.09]} & {\scriptsize [0.75, 3.13]} & \\
\addlinespace[2pt]
\multirow{2}{*}{TabPFN v3 $\rightarrow$ XGBoost} & \textbf{6.37} & 9.82 & 9.65 & 3.49 & \multirow{2}{*}{258/2/40} \\
 & {\scriptsize [5.33, 7.48]} & {\scriptsize [7.38, 12.94]} & {\scriptsize [6.61, 14.43]} & {\scriptsize [2.37, 4.91]} & \\
\bottomrule
\end{tabularx}
\vspace{3pt}
\end{table}

We use TALENT as an external validation benchmark to test whether the fixed
distillation recipe could be generalized.
We apply the recipe unchanged to all 300 TALENT datasets.
Across all four teacher--student pairs, KD improves over supevised trained
students on $236$--$258$ datasets,
reducing median primary error by $4.02$--$6.37\%$, with all paired 95\% confidence intervals above zero.

The improvement is not concentrated in a particular task.
Across binary classification, multiclass classification, and regression, all 12
teacher--student--task combinations show positive median improvements with
95\% confidence intervals above zero.
Together, these results show that the fixed recipe transfers across a substantially
broader collection of datasets without adapting new supervision
design.
\section{Conclusion}
We studied \emph{supervision construction} and \emph{query coverage} as two key design choices for distilling \acp{TFM} into lightweight, dataset-specific students.
Across all four teacher--student pairs, training solely on full-context teacher predictions achieves the best aggregate performance among the tested supervision configurations.
All four simple, task-agnostic query generators further improve aggregate performance for every pair, showing that useful teacher supervision can extend beyond observed training rows.

These findings yield a fixed recipe that trains students solely on full-context teacher predictions for observed and synthetic queries.
With default student configurations, this recipe outperforms architecture-matched tuned-and-ensembled students on TabArena and improves matched default students on TALENT without reselecting its target-generation settings.
Predictive gains vary across datasets and are associated with the teacher's advantage over the student trained directly on observed labels.
The distilled students can be served independently of the labeled context, trading a one-time distillation cost and some accuracy relative to their teachers for faster repeated inference, with the largest speedups in large-context settings.

\bibliography{iclr2027_conference}
\bibliographystyle{iclr2027_conference}
\appendix
\clearpage
\section{Experimental Details}
\subsection{Section 3 details}
\label{app:sec3-experimental-details}

\paragraph{Training protocol.}
We use TabArena pinned to commit \href{https://github.com/autogluon/tabarena/commit/eb5bc0816d4381f633beee17aa9bf028186e8983}{\texttt{eb5bc081}} (16 June 2026), and its 72-method official comparison preset.
Section 3 use TabArena splits 0--2 on 51 datasets with eight-fold student bagging.
Full-context targets use all fitting rows of each bagging member as teacher context; validation and test labels are excluded.
OOF targets use up to eight inner folds within that fitting set.

\paragraph{Implementation.}
We retain benchmark-default students and preprocessing, use eight teacher estimators, and apply no additional distillation temperature scaling.
Classification uses soft cross-entropy and regression uses squared error to teacher predictions.
TabM and XGBoost balance observed and synthetic queries through the sum of their group-mean losses.

\paragraph{Synthetic queries.}
The operator comparison uses one synthetic query per observed row.
CutMix replaces each feature with probability $0.5$; Mixup uses independent feature-wise $\operatorname{Unif}(0,1)$ coefficients.
Perturb uses numerical noise scaled by the column standard deviation and categorical replacement probability $0.5$; FullMix samples each feature independently from observed rows.
All source rows are sampled from the fitting set, and synthetic targets come only from the teacher.
The TabPFN v3--XGBoost budget comparison uses 152 matched dataset--split pairs because one $r=4$ result is unavailable; the other curves use 153.

\subsection{Section 4 details}
\label{app:sec4-experimental-details}

\paragraph{Evaluation protocol.}
TabArena evaluation covers all 816 official dataset--split pairs: nine splits for 34 datasets and 30 for 17, including design splits 0--2.
Appendix~\ref{app:post-design-evaluation} repeats the analyses in Sections~\ref{sec:eval_tabarena} and~\ref{sec:eval_regimes} after excluding design splits 0--2.
TALENT uses its fixed train/validation/test partitions and seeds 0--4, with training-only teacher context and no outer bagging or seed ensembling.
The released TALENT adapter uses the same \texttt{eb5bc081} dependency with a TabM compatibility layer.
The official tuned TabArena baselines select from one default and 200 random configurations; tuned ensembles use 40 ensemble-selection iterations.

\paragraph{Statistics.}
We average errors over matched splits or seeds before computing dataset-level relative gains, then report their median.
Paired bootstrap intervals resample datasets, using 20,000 replicates with joint recalibration of all 76 configurations for Section 4 Elo contrasts and 10,000 for TALENT median gains.
TALENT relative gains exclude reference errors at most $10^{-12}$, leaving 297 datasets; win/tie/loss counts use all 300, with absolute differences at most $10^{-12}$ treated as ties.
Each dataset receives equal weight.

\subsection{Timing details}
\label{app:section4-timing}

\paragraph{Serving measurements.}
We measure all 51 TabArena datasets at outer split 0 on one H200 host with 32 pinned CPU cores.
The reported latencies use resident models, 32-row requests, and teacher context recomputation without persistent KV caching.
After three warm-up calls, we record 7--30 GPU-synchronized calls, targeting two seconds of measured time, and take the median latency per dataset.
Initial model loading is excluded.

\paragraph{Cost aggregation.}
Offline cost includes teacher-target generation, student fitting, and training overhead.
Speedups and break-even request counts are computed per dataset before taking medians, retaining infinite break-even values when the student is no faster.
In Figure~\ref{fig:rq2_ratio}, generation time is the mean recorded cold target-generation cost across completed splits, excluding student fitting.

\section{Full Tabarena Results}
\label{app:full_results}

Table~\ref{tab:app-full-tabarena} reports all 72 official baseline configurations and our four KD students using default student configurations and $r=2$.
Elo, improvability, and win rate are computed jointly over exactly these 76 configurations.

\begingroup
\small
\setlength{\tabcolsep}{5pt}
\renewcommand{\arraystretch}{1.08}
\makeatletter
\setlength{\LTcapwidth}{\linewidth}
\@ifundefined{LT@\romannumeral\numexpr\c@LT@tables+1\relax}{}{%
  \setlength{\LTcapwidth}{0pt}%
  \let\app@LT@entry\LT@entry
  \def\LT@entry#1#2{\addtolength{\LTcapwidth}{#2}}%
  \csname LT@\romannumeral\numexpr\c@LT@tables+1\relax\endcsname
  \let\LT@entry\app@LT@entry
}
\makeatother
\begin{longtable}{@{}>{\raggedright\arraybackslash}p{0.57\linewidth}rrr@{}}
\caption{Full TabArena results on 51 datasets and 816 dataset--split pairs.
All metrics are computed jointly over the 76 displayed configurations: 72 official baselines and four KD students using default student configurations and $r=2$.
Shaded rows denote KD students.}%
\label{tab:app-full-tabarena}\\
\toprule
Model & Elo $\uparrow$ & \shortstack{Improvability\\(\%) $\downarrow$} & \shortstack{Win rate\\(\%) $\uparrow$} \\
\midrule
\endfirsthead
\multicolumn{4}{@{}l}{\textit{Table~\ref{tab:app-full-tabarena} continued.}}\\
\toprule
Model & Elo $\uparrow$ & \shortstack{Improvability\\(\%) $\downarrow$} & \shortstack{Win rate\\(\%) $\uparrow$} \\
\midrule
\endhead
\midrule
\multicolumn{4}{r@{}}{\textit{Continued on next page.}}\\
\endfoot
\bottomrule
\endlastfoot
% AutoGluon 1.5 (extreme, 4h)
AutoGluon 1.5 (extreme, 4h) & 1668.8 & 5.57 & 87.9 \\
% TA-TABPFN-3 (default)
TabPFN v3 (Default) & 1657.8 & 6.83 & 87.2 \\
% TABPFN-V2.6 (default)
TabPFN v2.6 (Default) & 1605.4 & 8.62 & 83.9 \\
% REALTABPFN-V2.5 (tuned + ensemble)
RealTabPFN v2.5 (Tuned+Ensemble) & 1585.5 & 8.26 & 82.4 \\
% TABICLV2 (default)
TabICL v2 (Default) & 1582.7 & 7.60 & 82.2 \\
% REALTABPFN-V2.5 (tuned)
RealTabPFN v2.5 (Tuned) & 1543.2 & 9.02 & 79.2 \\
% REALTABPFN-V2.5 (default)
RealTabPFN v2.5 (Default) & 1514.3 & 9.53 & 76.7 \\
% KD:TabPFN v3:60_tabm_cutmix_2_all_splits (default)
\rowcolor{blue!6}
KD: TabPFN v3 $\rightarrow$ TabM ($r=2$) & 1510.7 & 11.21 & 76.4 \\
% AutoGluon 1.4 (best, 4h)
AutoGluon 1.4 (best, 4h) & 1490.7 & 10.99 & 74.6 \\
% REALMLP_GPU (tuned + ensemble)
RealMLP (Tuned+Ensemble) & 1489.9 & 11.09 & 74.5 \\
% KD:TabICL v2:60_tabm_cutmix_2_all_splits (default)
\rowcolor{blue!6}
KD: TabICL v2 $\rightarrow$ TabM ($r=2$) & 1483.0 & 11.44 & 73.9 \\
% KD:TabPFN v3:61_xgboost_cutmix_2_all_splits (default)
\rowcolor{blue!6}
KD: TabPFN v3 $\rightarrow$ XGBoost ($r=2$) & 1457.8 & 12.41 & 71.4 \\
% TABDPT_GPU (tuned + ensemble)
TabDPT (Tuned+Ensemble) & 1443.2 & 11.58 & 70.0 \\
% KD:TabICL v2:61_xgboost_cutmix_2_all_splits (default)
\rowcolor{blue!6}
KD: TabICL v2 $\rightarrow$ XGBoost ($r=2$) & 1431.3 & 12.62 & 68.8 \\
% TABM_GPU (tuned + ensemble)
TabM (Tuned+Ensemble) & 1426.2 & 12.50 & 68.3 \\
% GBM (tuned + ensemble)
LightGBM (Tuned+Ensemble) & 1412.5 & 13.49 & 66.8 \\
% REALMLP_GPU (tuned)
RealMLP (Tuned) & 1412.2 & 12.47 & 66.8 \\
% CAT (tuned + ensemble)
CatBoost (Tuned+Ensemble) & 1398.1 & 13.10 & 65.3 \\
% TABDPT_GPU (tuned)
TabDPT (Tuned) & 1391.8 & 12.81 & 64.6 \\
% TA-ILTM (tuned + ensemble)
iLTM (Tuned+Ensemble) & 1391.7 & 13.56 & 64.6 \\
% CAT (tuned)
CatBoost (Tuned) & 1388.7 & 13.30 & 64.3 \\
% MNCA_GPU (tuned + ensemble)
ModernNCA (Tuned+Ensemble) & 1375.3 & 13.78 & 62.8 \\
% TABM_GPU (tuned)
TabM (Tuned) & 1373.7 & 13.36 & 62.7 \\
% GBM (tuned)
LightGBM (Tuned) & 1370.2 & 14.16 & 62.3 \\
% XGB (tuned + ensemble)
XGBoost (Tuned+Ensemble) & 1359.9 & 14.26 & 61.2 \\
% CAT (default)
CatBoost (Default) & 1353.2 & 14.06 & 60.4 \\
% TA-LIMIX (default)
LimiX (Default) & 1350.8 & 13.57 & 60.2 \\
% MNCA_GPU (tuned)
ModernNCA (Tuned) & 1344.2 & 14.30 & 59.4 \\
% XRFM_GPU (tuned + ensemble)
xRFM (Tuned+Ensemble) & 1339.7 & 14.67 & 58.9 \\
% XGB (tuned)
XGBoost (Tuned) & 1338.1 & 14.57 & 58.7 \\
% TABPFNV2_GPU (tuned + ensemble)
TabPFN v2 (Tuned+Ensemble) & 1331.5 & 14.92 & 58.0 \\
% MITRA_GPU (default)
Mitra (Default) & 1318.7 & 15.43 & 56.6 \\
% TABDPT_GPU (default)
TabDPT (Default) & 1318.0 & 15.19 & 56.5 \\
% TABICL_GPU (default)
TabICL (Default) & 1309.3 & 15.06 & 55.5 \\
% XRFM_GPU (tuned)
xRFM (Tuned) & 1294.7 & 15.89 & 53.8 \\
% TA-ILTM (tuned)
iLTM (Tuned) & 1288.4 & 15.47 & 53.1 \\
% TABM_GPU (default)
TabM (Default) & 1285.2 & 15.57 & 52.7 \\
% NN_TORCH (tuned + ensemble)
TorchMLP (Tuned+Ensemble) & 1275.8 & 15.48 & 51.7 \\
% TABPFNV2_GPU (tuned)
TabPFN v2 (Tuned) & 1275.2 & 16.46 & 51.6 \\
% SAP-RPT-OSS (default)
SAP-RPT-OSS (Default) & 1273.2 & 16.60 & 51.4 \\
% BETA_GPU (default)
BetaTabPFN (Default) & 1270.8 & 17.09 & 51.1 \\
% EBM (tuned + ensemble)
EBM (Tuned+Ensemble) & 1256.5 & 17.28 & 49.5 \\
% TABPFNV2_GPU (default)
TabPFN v2 (Default) & 1248.1 & 17.23 & 48.5 \\
% MNCA_GPU (default)
ModernNCA (Default) & 1242.5 & 17.61 & 47.9 \\
% REALMLP_GPU (default)
RealMLP (Default) & 1223.1 & 16.97 & 45.7 \\
% EBM (tuned)
EBM (Tuned) & 1222.9 & 17.94 & 45.7 \\
% XGB (default)
XGBoost (Default) & 1209.8 & 17.35 & 44.2 \\
% NN_TORCH (tuned)
TorchMLP (Tuned) & 1205.2 & 17.25 & 43.7 \\
% XT (tuned + ensemble)
ExtraTrees (Tuned+Ensemble) & 1199.0 & 18.61 & 43.0 \\
% FASTAI (tuned + ensemble)
FastaiMLP (Tuned+Ensemble) & 1196.5 & 18.10 & 42.7 \\
% EBM (default)
EBM (Default) & 1192.5 & 18.96 & 42.3 \\
% GBM (default)
LightGBM (Default) & 1183.1 & 17.91 & 41.2 \\
% RF (tuned + ensemble)
RandomForest (Tuned+Ensemble) & 1174.5 & 19.54 & 40.3 \\
% XT (tuned)
ExtraTrees (Tuned) & 1165.7 & 19.53 & 39.3 \\
% FASTAI (tuned)
FastaiMLP (Tuned) & 1138.7 & 19.55 & 36.4 \\
% RF (tuned)
RandomForest (Tuned) & 1137.3 & 20.28 & 36.3 \\
% PB (tuned + ensemble)
PerpetualBooster (Tuned+Ensemble) & 1089.2 & 25.05 & 31.3 \\
% TA-ORION-MSP (default)
OrionMSP (Default) & 1088.7 & 22.02 & 31.3 \\
% TA-ILTM (default)
iLTM (Default) & 1088.2 & 21.76 & 31.2 \\
% TABSTAR (tuned)
TabSTAR (Tuned) & 1088.1 & 24.47 & 31.2 \\
% TABSTAR (tuned + ensemble)
TabSTAR (Tuned+Ensemble) & 1087.2 & 24.49 & 31.1 \\
% NN_TORCH (default)
TorchMLP (Default) & 1073.8 & 21.39 & 29.8 \\
% PB (tuned)
PerpetualBooster (Tuned) & 1051.6 & 26.44 & 27.7 \\
% XRFM_GPU (default)
xRFM (Default) & 1041.5 & 25.05 & 26.8 \\
% TABFLEX_GPU (default)
TabFlex (Default) & 1010.8 & 26.42 & 24.1 \\
% FASTAI (default)
FastaiMLP (Default) & 1004.7 & 24.27 & 23.5 \\
% RF (default)
RandomForest (Default) & 1000.0 & 25.08 & 23.1 \\
% KNN (tuned + ensemble)
KNN (Tuned+Ensemble) & 992.4 & 26.76 & 22.5 \\
% TABSTAR (default)
TabSTAR (Default) & 988.4 & 29.50 & 22.2 \\
% XT (default)
ExtraTrees (Default) & 982.1 & 26.47 & 21.7 \\
% LR (tuned + ensemble)
Linear (Tuned+Ensemble) & 956.6 & 32.71 & 19.7 \\
% LR (tuned)
Linear (Tuned) & 931.9 & 33.27 & 17.9 \\
% PB (default)
PerpetualBooster (Default) & 930.7 & 30.57 & 17.8 \\
% KNN (tuned)
KNN (Tuned) & 886.0 & 31.87 & 14.8 \\
% LR (default)
Linear (Default) & 856.1 & 35.92 & 13.0 \\
% KNN (default)
KNN (Default) & 643.7 & 45.28 & 4.4 \\
\end{longtable}
\endgroup

\clearpage
\section{Section 4 Results Excluding Design Splits}
\label{app:post-design-evaluation}

We repeat the TabArena analyses in Sections~\ref{sec:eval_tabarena} and~\ref{sec:eval_regimes} after excluding outer splits 0--2, which were used for the design studies in Section~\ref{sec:design}.
This leaves 663 dataset--split pairs across all 51 datasets: splits 3--8 for 34 datasets and splits 3--29 for 17 datasets.
We retain the fixed full-context, teacher-only CutMix recipe with default student configurations and $r=2$, without reselecting any design choice.
This analysis checks sensitivity to reusing the design splits; the datasets themselves remain the same, so it does not constitute evaluation on unseen datasets.

\paragraph{Aggregation and uncertainty.}
Elo, improvability, and win rate are recomputed jointly over the same 72 official baseline configurations and four KD configurations, with equal dataset weights.
For relative-error analyses, we first average errors over the retained splits within each dataset, then compute dataset-level ratios and report their medians.
All intervals below use 20,000 paired dataset-bootstrap resamples and the 2.5th and 97.5th percentiles; Elo intervals jointly refit all 76 configurations in each resample.
Win/tie/loss counts use dataset-mean errors and an absolute tie tolerance of $10^{-12}$.

\subsection{Comparison with Supervised Students}
\label{app:post-design-baselines}

The conclusion of Section~\ref{sec:eval_tabarena} holds after excluding the design splits.
All four KD configurations outperform the matched tuned-and-ensembled student, with gains of $56.4$--$101.6$ Elo and median relative-error reductions of $1.37$--$1.84\%$ (Table~\ref{tab:app-post-design}).
Every paired interval is above zero, and KD wins on $33$--$36$ of the 51 datasets.

% Generated by scripts/build_post_design_section4.py; splits >= 3 only.
\begin{table}[!htbp]
\centering
\small
\caption{\textbf{KD versus tuned-and-ensembled students after excluding design splits.}
Results use 51 datasets and 663 dataset--split pairs, omitting splits 0--2.
KD gain is the median relative primary-error reduction against the matched tuned ensemble; W/T/L counts dataset-level wins, ties, and losses.
Elo is jointly recalibrated over all 76 configurations.
Brackets show paired dataset-bootstrap 95\% intervals (20,000 resamples).}
\label{tab:app-post-design}
\setlength{\tabcolsep}{5pt}
\renewcommand{\arraystretch}{1.08}
\begin{tabularx}{\linewidth}{@{}Xrrr@{}}
\toprule
Model & $\Delta$Elo $\uparrow$ & KD gain (\%) $\uparrow$ & W/T/L \\
\midrule
TabICL v2 $\rightarrow$ TabM & +56.4 [4.2, 106.6] & 1.37 [0.45, 2.03] & 34/0/17 \\
TabICL v2 $\rightarrow$ XGBoost & +72.1 [14.3, 130.3] & 1.64 [0.50, 2.34] & 33/0/18 \\
\addlinespace[2pt]
TabPFN v3 $\rightarrow$ TabM & +86.0 [34.0, 138.5] & 1.55 [0.71, 2.28] & 35/0/16 \\
TabPFN v3 $\rightarrow$ XGBoost & +101.6 [44.3, 162.0] & 1.84 [0.87, 3.78] & 36/0/15 \\
\bottomrule
\end{tabularx}
\end{table}

Table~\ref{tab:app-post-design-supervised} repeats the broader comparison from Table~\ref{tab:section4-tabarena-supervised} on the retained splits.
Each KD configuration exceeds the highest Elo among the configurations of at least 14 of the 15 conventional supervised families.
TabPFN v3--TabM exceeds all 15; RealMLP remains the only family whose strongest configuration ranks above the other three KD models.

% Generated by scripts/build_post_design_section4.py; splits >= 3 only.
\begin{table}[!htbp]
\centering
\small
\caption{\textbf{Broader TabArena comparison after excluding design splits.}
The displayed models match Table~\ref{tab:section4-tabarena-supervised}; all metrics use the 76-configuration pool on the remaining 663 dataset--split pairs.
KD students use default configurations and $r=2$.
Shading marks KD students; \textcolor{Red}{\textbf{1st}}, \textcolor{Orange}{\textbf{2nd}}, and \textcolor{Emerald}{\textbf{3rd}} results among displayed non-teacher models are highlighted.}
\label{tab:app-post-design-supervised}
\setlength{\tabcolsep}{5pt}
\renewcommand{\arraystretch}{1.08}
\begin{tabularx}{\linewidth}{@{}Xrrr@{}}
\toprule
Model & Elo $\uparrow$ & Improvability (\%) $\downarrow$ & Win rate (\%) $\uparrow$ \\
\midrule
\rowcolor{blue!6}
TabPFN v3 $\rightarrow$ TabM & \textcolor{Red}{\textbf{1513.5}} & \textcolor{Orange}{\textbf{11.18}} & \textcolor{Red}{\textbf{76.5}} \\
RealMLP (Tuned+Ensemble) & \textcolor{Orange}{\textbf{1490.8}} & \textcolor{Red}{\textbf{11.11}} & \textcolor{Orange}{\textbf{74.5}} \\
\rowcolor{blue!6}
TabICL v2 $\rightarrow$ TabM & \textcolor{Emerald}{\textbf{1484.0}} & \textcolor{Emerald}{\textbf{11.43}} & \textcolor{Emerald}{\textbf{73.9}} \\
\rowcolor{blue!6}
TabPFN v3 $\rightarrow$ XGBoost & 1460.6 & 12.36 & 71.6 \\
\rowcolor{blue!6}
TabICL v2 $\rightarrow$ XGBoost & 1431.1 & 12.63 & 68.6 \\
TabM (Tuned+Ensemble) & 1427.6 & 12.49 & 68.3 \\
LightGBM (Tuned+Ensemble) & 1412.5 & 13.48 & 66.7 \\
CatBoost (Tuned+Ensemble) & 1401.2 & 13.02 & 65.5 \\
TabM (Tuned) & 1373.9 & 13.39 & 62.6 \\
XGBoost (Tuned+Ensemble) & 1359.0 & 14.27 & 60.9 \\
XGBoost (Tuned) & 1337.5 & 14.59 & 58.5 \\
TabM (Default) & 1284.9 & 15.65 & 52.5 \\
XGBoost (Default) & 1208.5 & 17.41 & 43.9 \\
\midrule
\multicolumn{4}{@{}l}{\textit{Teacher models}} \\
\addlinespace[2pt]
TabPFN v3 & 1659.2 & 6.85 & 87.3 \\
TabICL v2 & 1581.9 & 7.53 & 82.1 \\
\bottomrule
\end{tabularx}
\end{table}

\FloatBarrier

\subsection{Teacher Advantage and KD Gain}
\label{app:post-design-transfer}

\begin{figure}[!htbp]
\centering
\includegraphics[width=\linewidth]{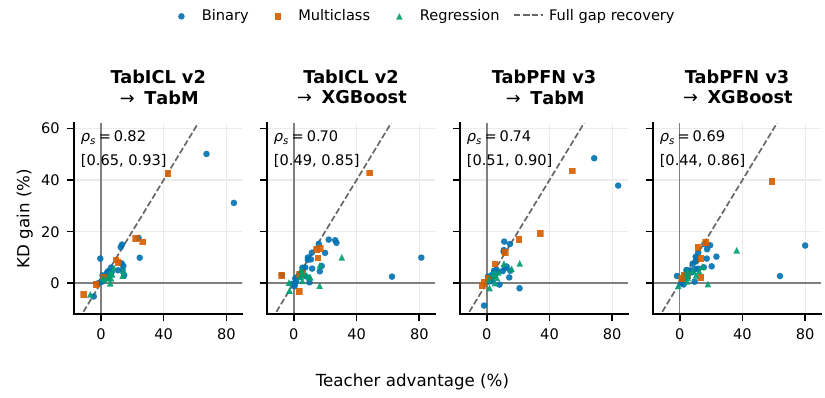}
\caption{\textbf{Teacher advantage and KD gain after excluding design splits.}
Each point is one of the 51 TabArena datasets, using errors averaged over splits $\geq 3$.
Teacher and KD gains are relative error reductions over the default student, $100(1-E/E_{\rm raw})$; the dashed line denotes full recovery.
Annotations report Spearman correlations with paired dataset-bootstrap 95\% intervals (20,000 resamples).
These correlations are descriptive because both axes share the raw baseline.}
\label{fig:app-post-design-transfer}
\end{figure}

The relationship between teacher advantage and KD gain in Section~\ref{sec:eval_regimes} also persists.
Across all 51 datasets, Spearman correlations range from $0.69$ to $0.82$, with all bootstrap intervals above zero (Figure~\ref{fig:app-post-design-transfer}).
The teacher and KD gains share the raw-student denominator, so these correlations remain descriptive.

On the $42$--$48$ datasets per pair where the teacher reduces raw-student error by more than $1\%$, KD recovers a median $67.8$--$73.4\%$ of the teacher--student error gap (Table~\ref{tab:app-post-design-transfer}).
The eligible datasets are selected again using only the retained splits, and recovery intervals resample these eligible datasets.
KD still loses to the raw student on one to three eligible datasets per pair, so a stronger teacher does not ensure improvement on every dataset.

% Generated by scripts/build_post_design_section4.py; splits >= 3 only.
\begin{table}[!htbp]
\centering
\small
\caption{\textbf{Recovery of teacher gains after excluding design splits.}
Errors are averaged over retained splits before selecting datasets where the teacher reduces default-student error by more than $1\%$.
Gap recovery is $100(E_{\rm raw}-E_{\rm KD})/(E_{\rm raw}-E_T)$.
Entries report the median with dataset-bootstrap 95\% intervals (20,000 resamples); W/T/L compares KD with the raw student on these $n$ datasets.}
\label{tab:app-post-design-transfer}
\setlength{\tabcolsep}{5pt}
\renewcommand{\arraystretch}{1.08}
\begin{tabularx}{\linewidth}{@{}Xrrr@{}}
\toprule
Model & $n$ & Gap recovery (\%) & KD W/T/L \\
\midrule
TabICL v2 $\rightarrow$ TabM & 42 & 71.8 [56.0, 89.8] & 41/0/1 \\
TabICL v2 $\rightarrow$ XGBoost & 43 & 69.0 [53.9, 83.1] & 41/0/2 \\
\addlinespace[2pt]
TabPFN v3 $\rightarrow$ TabM & 45 & 73.4 [56.4, 86.2] & 42/0/3 \\
TabPFN v3 $\rightarrow$ XGBoost & 48 & 67.8 [52.7, 78.8] & 46/0/2 \\
\bottomrule
\end{tabularx}
\end{table}

\FloatBarrier

\end{document}